\documentclass[runningheads]{llncs}
\usepackage{graphicx}
\usepackage{comment}
\usepackage{amsmath,amssymb} % define this before the line numbering.
\usepackage{color}

\usepackage{multirow}
\usepackage{multicol}
\usepackage{floatrow}
\usepackage[noend]{algpseudocode}
\usepackage{algorithmicx,algorithm}
\usepackage{bbding}
\usepackage[breaklinks=true,bookmarks=false]{hyperref}

\begin{document}
\pagestyle{headings}
\mainmatter
\def\ECCVSubNumber{566}  % Insert your submission number here
\newfloatcommand{capbtabbox}{table}[][\FBwidth]

\title{Faster Person Re-Identification} % Replace with your title

% % % INITIAL SUBMISSION 

% %\begin{comment}
% \titlerunning{ECCV-20 submission ID \ECCVSubNumber} 
% \authorrunning{ECCV-20 submission ID \ECCVSubNumber} 
% \author{Anonymous ECCV submission}
% \institute{Paper ID \ECCVSubNumber}
% %\end{comment}
% %******************

\author{Guan'an Wang\inst{1,3}\thanks{\scriptsize This work was done when Guan'an Wang was at QMUL supervised by \textit{Prof.} Shaogang Gong} \and
Shaogang Gong\inst{2} \and
Jian Cheng\inst{3,4,5}\and
Zengguang Hou\inst{1,3,5}}

%
%Please write out author names in full in the paper, i.e. full given and family names.
%If any authors have names that can be parsed into FirstName LastName in multiple ways, please include the correct parsing, in a comment to the volume editors:
%\index{Lastnames, Firstnames}
%(Do not uncomment it, because you may introduce extra index items if you do that, we will use scripts for introducing index entries...)
\authorrunning{Wang et al.}
% Replace with shorter version of the author list. If there are more authors than fits a line, please use A. Author et al.
%

% \institute{$^1$SKLMCCS, CASIA \ \ $^2$SAI, UCAS \ \  $^3$QMUL \ \  $^4$NLPR, CASIA \\
% \email{wangguanan2015@ia.ac.cn, s.gong@qmul.ac.uk, \\ jcheng@nlpr.ia.ac.cn, zengguang.hou@ia.ac.cn}
% }

\institute{
The State Key Laboratory of Management and Control of Complex System, Institute of Automation, Chinese Academy of Sciences (CAS). \and
Queen Mary University of London. \and
School of Artificial Intelligence, University of Chinese Academy of Sciences. \and
National Laboratory of Pattern Recognition, Institute of Automation, CAS. \and
CAS Center for Excellence in Brain Science and Intelligence Technology. 
\email{wangguanan2015@ia.ac.cn, s.gong@qmul.ac.uk, \\ jcheng@nlpr.ia.ac.cn, zengguang.hou@ia.ac.cn}
}

% % CAMERA READY SUBMISSION
% \begin{comment}
% % If the paper title is too long for the running head, you can set
% % an abbreviated paper title here
% \end{comment}
%******************
\maketitle

\begin{abstract}

Fast person re-identification (ReID) aims to search person images quickly and accurately.
The main idea of recent fast ReID methods is the hashing algorithm,
which learns compact binary codes and performs fast Hamming distance
and counting sort. However, a very long code is
needed for high accuracy (\textit{e.g.} 2048), which compromises search speed.
In this work, we introduce a new solution for fast ReID by formulating a
novel Coarse-to-Fine (CtF) hashing code search strategy, which complementarily uses
short and long codes, achieving both faster speed and better accuracy. It
uses shorter codes to coarsely rank broad matching similarities and longer codes
to refine only a few top candidates for more accurate instance ReID. 
Specifically, we design an All-in-One (AiO) framework together with a
Distance Threshold Optimization (DTO) algorithm. 
In AiO, we simultaneously learn and enhance multiple codes of
different lengths in a single model. It learns multiple codes in a
pyramid structure, and encourage shorter codes to mimic longer codes
by self-distillation. 
DTO solves a complex threshold search problem by a simple optimization
process, and the balance between accuracy and speed is easily
controlled by a single parameter. 
It formulates the optimization target as a $F_{\beta}$ score that can
be optimised by Gaussian cumulative distribution functions.
Experimental results on 2 datasets show that our proposed method (CtF) is
not only $8\%$ more accurate but also $5\times$ faster than
contemporary hashing ReID methods. Compared with non-hashing ReID
methods, CtF is $50\times$ faster with comparable accuracy.
Code is available at \url{https://github.com/wangguanan/light-reid}.

\end{abstract}

\section{Introduction}

Person re-identification (ReID) \cite{gong2014person,zheng2016person}
aims to match images of a person across disjoint cameras, which is
widely used in video surveillance, security and smart city. 
Many methods \cite{ma2014covariance,yang2014salient,liao2015person,zheng2013reidentification,koestinger2012large,liao2015efficient,zheng2016person,hermans2017defense,sun2018beyond} have been proposed for person ReID.
However, for higher accuracy, most of them utilize a large deep network
to learn high-dimensional real-value features for computing
similarities by Euclidean distance and returning a rank list by
quick-sort \cite{hoare1962quicksort}. Quick-sort of high-dimensional deep features can be
slow, especially when the gallery set is large. 
Table \ref{table:sort_time} shows that the query time per ReID probe
image increases massively with the increase of the ReID gallery size;
and counting-sort \cite{bajpai2014implementing} is much more efficient than quick-sort, in which the
former has a linear complexity w.r.t the gallery size ($O(n)$) whilst
the latter has a logarithm complexity ($O(nlogn)$).

Several fast ReID methods
\cite{chen2016person,zheng2016learning,wu2017structured,zhu2017part,chen2017fast,fang2018perceptual,zhu2018fast,liu2019adversarial}
have been proposed to increase ReID speed whist retaining ReID accuracy.
The common main idea is hashing, which learns a binary code
instead of real-value features. To sort binary codes, the
inefficient Euclidean distance and quick-sort are replaced by
the Hamming-distance and counting-sort \cite{bajpai2014implementing}. 
Table \ref{table:distance_time} shows that computing a
Hamming distance between 2048-dimensional binary-codes is $229\times$
faster than that of a Euclidean distance between real-value features. 

\begin{table*}[t]
\begin{floatrow}
\capbtabbox{
\begin{tabular}{c|cc}
\hline
\hline
\multirow{1}{*}{Gallery} & \multicolumn{2}{c}{Query Time (s)} \\
\multirow{1}{*}{Size} & \multicolumn{1}{c}{ \ \ Quick-Sort \ \ } & \multicolumn{1}{c}{Counting-Sort}\\
\hline
% $1 \times 10^2$  &   $2.0 \times 10^{-4}$ &   $1.6 \times 10^{-4}$ \\
$1 \times 10^3$  &   $3.4 \times 10^{-3}$       &   $4.7 \times 10^{-4}$ \\
$1 \times 10^4$  &   $1.0 \times 10^{-1}$       &   $2.7 \times 10^{-3}$ \\
$1 \times 10^5$  &   $4.3 \times 10^{-1}$       &   $2.7 \times 10^{-2}$ \\
$1 \times 10^6$  &   $6.4 \times 10^{0}$ \ \    &   $2.6 \times 10^{-1}$ \\
$1 \times 10^7$  &   $1.1 \times 10^{2}$ \ \    &   $2.7 \times 10^{0}$ \ \  \\
\hline
Per Sample & -          & $2.6 \times 10^{-7}$ \\
Complexity & $O(nlogn)$ & $O(n)$ \\
\hline
\hline
\end{tabular}
}{
\caption{ReID search time per probe image by quick-sort (real-value) and
  counting-sort (binary). The latter is much faster.}
\label{table:sort_time}
}

\capbtabbox{
\begin{tabular}{c|cc}
\hline
\hline
\multirow{1}{*}{Code} & \multicolumn{2}{c}{Computation Time (s)} \\
\multirow{1}{*}{Length} & \ \ Euclidean \ \  & \ \ Hamming \ \  \\
\hline
$32$    &   $6.8 \times 10^{-5}$ &   $2.4 \times 10^{-6}$ \\
$64$    &   $1.3 \times 10^{-4}$ &   $2.7 \times 10^{-6}$ \\
$128$   &   $2.6 \times 10^{-4}$ &   $2.8 \times 10^{-6}$ \\
$256$   &   $5.0 \times 10^{-4}$ &   $3.3 \times 10^{-6}$ \\
$512$   &   $1.0 \times 10^{-3}$ &   $4.4 \times 10^{-6}$ \\
$1,024$  &   $2.0 \times 10^{-3}$ &   $7.1 \times 10^{-6}$ \\
$2,048$  &   $3.9 \times 10^{-3}$ &   $1.7 \times 10^{-5}$ \\
\hline
\hline
\end{tabular}
}{
\caption{Comparing Euclidean and Hamming
  distances, Euclidean and longer lengths are slow to compute.}
\label{table:distance_time}
}

\end{floatrow}
\end{table*}

Different from common image retrieval tasks, which are category-level
matching in a close-set, ReID is instance-level matching in an open-set (zero-shot setting). 
For image retrieval in the ImageNet \cite{russakovsky2015imagenet},
the classes of training and test sets are the same and imagery
appearances of different classes diverse a lot, such as dog, car, and airplane.
In contrast, the training and test ReID images have completely
different ID classes without any overlap (ZSL) whilst the appearances of
different persons can be very similar to subtle changes
(fine-grained) on clothing,
body characteristics, gender, and carried-objects.
The ZSL and fine-grained characteristics of ReID require
state-of-the-art hashing-based fast ReID
models \cite{liu2019adversarial} to employ very long binary codes,
\textit{e.g.} 2048, in order to retain competitive ReID accuracy.
However, the binary code length affects significantly
the cost of computing Hamming distance. Table
\ref{table:distance_time} shows that computing a Hamming distance
between two 2048-dimensional binary codes takes $1.7\times10^{-5}$
seconds, which is $7\times$ slower than computing that of
32-dimensional binary codes at $2.4\times10^{-6}$ seconds. 
This motivates us to solve the following problem: How to yield
higher accuracy from hashing-based ReID using shorter binary codes.

\begin{figure}[t]
\center
\includegraphics[scale=0.4]{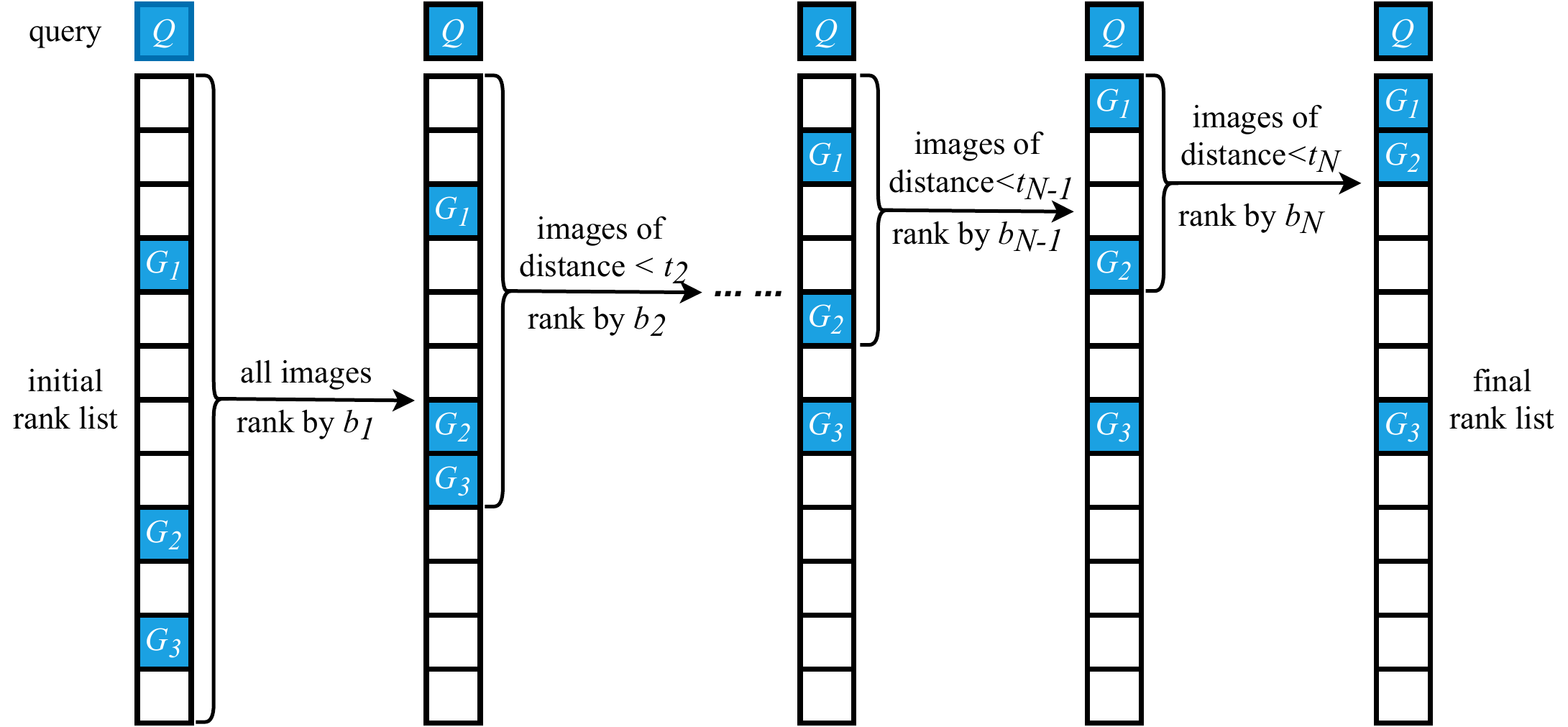}
\caption{A Coarse-to-Fine (CtF) hashing code search strategy to speed up ReID,
  where $Q$ is a query image, $\{G_i\}_{i=1}^{3}$ are the positive images in the gallery set, $B=\{b_k\}_{k=1}^{N}$ are binary codes of
  lengths $L=\{l_k\}_{k=1}^{N}$, $T=\{k_i\}_{k=2}^{N}$ are
  Hamming distance thresholds where gallery images are selected by
  each $t_k$ for further comparison by increasingly longer codes $b_{k}$.
}
\label{figure:coarse-to-fine strategy}
\end{figure}

To that end, we propose a novel Coarse-to-Fine (CtF) search strategy
for faster ReID whilst also retaining competitive accuracy. 
At test time, our model (CtF) first uses shorter codes to coarsely
rank a gallery, then iteratively utilises longer codes to further rank
selected top candidates where the top-ranked candidates are defined
iteratively by a set of Hamming distance thresholds. Thus, the long
codes are only used for a decreasingly fewer matches in ranking in
order to reduce the overall search time whilst retaining ReID accuracy.
This is an intuitively straightforward idea but not easily
computable for ReID due to three difficulties:
(1) Coarse-to-fine search requires multiple codes of different
lengths. Asymmetrically, computing them with multiple models is
both time-consuming and sub-optimal. 
(2) The coarse ranking must be accurate enough to minimise missing
true-match candidates in fine-grained ranking whilst keeping
their numbers small, thus reduce the total search time. Paradoxically,
shorter codes perform much worse than longer codes in ReID task
therefore hard to be sufficiently accurate. 
(3) The set of distance thresholds for guiding the coarse search affect both final accuracy and overall speed. How to determine {\em automatically} these
thresholds to balance optimally accuracy and speed is both
important and nontrivial.

In this work, we propose a novel All-in-One (AiO) framework together
with a Distance Threshold Optimization (DTO) algorithm to
simultaneously solve these three problems.
The AiO framework can simultaneously learn and enhance multiple codes
of different lengths in a single model. It progressively learns
multiple codes in a pyramid structure, where the knowledge from the
bottom long code is shared by the top short code. We
promote shorter codes to mimic longer codes by both probability- and
similarity- distillation. This makes shorter codes more powerful
without importing extra teacher networks.
The DTO algorithm solves a complex threshold search problem by a
simple optimization process and the balance between search accuracy and speed
is easily controlled by a single parameter. 
It explores a $F_{\beta}$ score as the optimization target formulated as
Gaussian cumulative distribution functions. So that we can estimate its
parameters by the statistics of Gaussian probability distributions
modeling the distances of positive and negative pairs. Finally, by
maximizing the $F_{\beta}$ score, we compute iteratively optimal distance
thresholds.

Our contributions are:
(1) We propose a novel Coarse-to-Fine (CtF) search strategy for Faster
ReID, not only speeding up hashing ReID, but also improving their accuracy. To the best of our knowledge, this is the first work to introduce such search strategy into ReID.
(2) A novel All-in-One (AiO) framework is proposed to learn and enhance multiple codes of different lengths in a single framework by viewing it as a multi-channel self-distillation problem. In the framework, the multiple codes are learned in a pyramid structure and shorter codes mimic longer codes via probability- and similarity- distillation loss.
(3) A novel Distance Threshold Optimization (DTO) algorithm is proposed to find the optimal thresholds for coarse-to-fine search by concluding the threshold search task to a $F_{\beta}$ distance optimization problem. The $F_{\beta}$ score is represented with Gaussian cumulative distribution functions, whose mean and variance can be estimated by fitting a small validation set.
(4) Extensive experimental results on two datasets show that, our proposed method is $50\times$ faster than non-hashing ReID methods, $5\times$ faster and $8\%$ more accurate than hashing ReID methods.

\section{Related Works}

In this work, we wish to solve the fast ReID problem under the framework of hashing by proposing an All-in-One (AiO) hashing learning module and a Distance Threshold Optimization (DTO) algorithm. Thus, we mainly discuss the related works including non-fast person re-identification (ReID) task, fast ReID task and hashing algorithm. 

\textbf{Person Re-Identification.}
Person re-identification addresses the problem of matching pedestrian images across disjoint cameras \cite{gong2014person}. 
The key challenges lie in the large intra-class and small inter-class variation caused by different views, poses, illuminations, and occlusions.
Existing methods can be grouped into hand-crafted descriptors \cite{ma2014covariance,yang2014salient,liao2015person}, metric learning methods \cite{zheng2013reidentification,koestinger2012large,liao2015efficient} and deep learning algorithms \cite{zheng2016person,hermans2017defense,sun2018beyond,wang2019rgb,wang2019color,wang2020cross,wang2020high}. 
The goal of hand-crafted descriptors is to design robust features.
Metric learning aims to make a pair of true matches have a relatively smaller distance than that of a wrong match pair in a discriminant manner.
Deep learning algorithms adopt deep neural networks to straightly learn robust and discriminative features in an end-to-end manner and achieve the best performance. 
%
% Here, we mainly show some deep learning methods.
%
% For example, Zheng \textit{et al.} \cite{zheng2016person} learn identity-discriminative features by fine-tuning a pre-trained CNN to minimize a classification loss.  
% In \cite{hermans2017defense}, Hermans \textit{et al.} show that using a variant of the triplet loss outperforms most other published methods by a large margin.
% In \cite{sun2018beyond}, a network named Part-based Convolutional Baseline (PCB) is proposed to learn fine-grained part-level features with a uniform partition strategy.
%
However, all the ReID methods above learn real-value features for high accuracy, which is slow.

\textbf{Hashing Algorithm.}
Hashing algorithm mainly divided into unsupervised and (semi-)supervised ones.
Unsupervised hashing methods (LSH \cite{Datar2004Locality}, SH \cite{Weiss2008Spectral}, ITQ \cite{Lazebnik2011Iterative}) employ unlabeled data even no data.
(Semi-)Supervised (SSH \cite{Wang2012Semi}, BRE \cite{Kulis2009Learning}, KSH \cite{Liu2012Supervised}, SDH \cite{Shen2015Supervised}, SSGAH\cite{wang2018semi}) utilize labeled information to improve binary codes.
Recently, inspired by powerful deep networks, some deep hashing methods (CNNH \cite{Xia2014Supervised}, NINH \cite{Lai2015Simultaneous}, DPSH \cite{Li2016Feature}) have been proposed and achieve much better performance. They usually utilize a CNN to extract meaningful features, formulate the hashing function as a fully-connected layer with $tanh/sigmoid$ activation function, and quantize features by $signature$ function. The framework can be optimized with a related layer or some iteration strategies.
However, all the hashing methods are designed for close-set category-level retrieval tasks, which cannot be directly used for person ReID, an open-set fine-grained search problem.

\textbf{Fast Person Re-Identification.}
Fast ReID methods aims to search in a fast speed meanwhile obtaining accuracy as high as possible.
The main idea of those methods is hashing algorithm, which learns binary code instead of real-value features. Based on the binary codes, the inefficient Euclidean distance and quick-sorting can be replaced by efficient Hamming distance and counting sort.
Zheng \textit{et al.} \cite{zheng2016learning} learn cross-view binary codes using two hash functions for two different views.
Wu \textit{et al.} \cite{wu2017structured} simultaneously learn both CNN feature and hash functions to get robust yet discriminative features and similarity-preserving binary codes.
CSBT \cite{chen2017fast} solves the cross-camera variations problem by employing a subspace projection to maximize intra-person similarity and inter-person discrepancies.
In \cite{zhu2017part} integrate spatial information for discriminative features by representing horizontal parts to binary codes.
ABC \cite{liu2019adversarial} improves binary codes by implicitly fits the feature distribution to a pre-defined binary one with Wasserstein distance.
However, all the fast ReID methods take very long binary codes (\textit{e.g.} 2048) for high accuracy.
Different from them, we propose a coarse-to-fine search strategy which complementarily uses codes of different lengths, obtaining not only faster speed but also higher accuracy.

\section{Proposed Method}

In this work, we propose a coarse-to-fine (CtF) search strategy for fast and accurate ReID.
For effectively implementing the strategy, we design an All-in-One (AiO) framework together with a Distance Threshold Optimization (DTO) algorithm.
The former learns and enhances multiple codes of different lengths in a single framework.
The latter finds the optimal distance thresholds to balance time and accuracy.

\subsection{Coarse-to-Fine Search}

As we illustrated in the introduction section, although the long binary codes can get high accuracy, it takes much longer time than short codes. This motivates us to think about that can we reduce the usage of long codes to further speed hashing ReID methods up.
Thus, a simple but efficient solution is complementarily using both short and long codes. Here, shorter codes fast return a rough rank list of gallery, and longer codes carefully refine a small number of top candidates.
Figure  \ref{figure:coarse-to-fine strategy} shows its procedures.

Although the idea is straightforward, there are still three difficulties preventing it being applied to ReID.
(1) Coarse-to-fine search requires multiple codes of different
lengths. Asymmetrically, computing them with multiple models is
both time-consuming and sub-optimal. 
(2) The coarse ranking must be accurate enough to minimise missing
true-match candidates in fine-grained ranking whilst keeping
their numbers small, thus reduce the total search time. Paradoxically,
shorter codes perform much worse than longer codes in ReID task.
(3) The set of distance thresholds for guiding the coarse search affect both final accuracy and overall speed. How to determine {\em automatically} these
thresholds to balance optimally accuracy and speed is both
important and nontrivial.
To solve the problems, we propose an All-in-One (AiO) framework and a Distance Threshold Optimization (DTO) algorithm. Please see the next two parts for more details.

\subsection{All-in-One Framework}

\begin{figure}[t]
\center
\includegraphics[width=\linewidth]{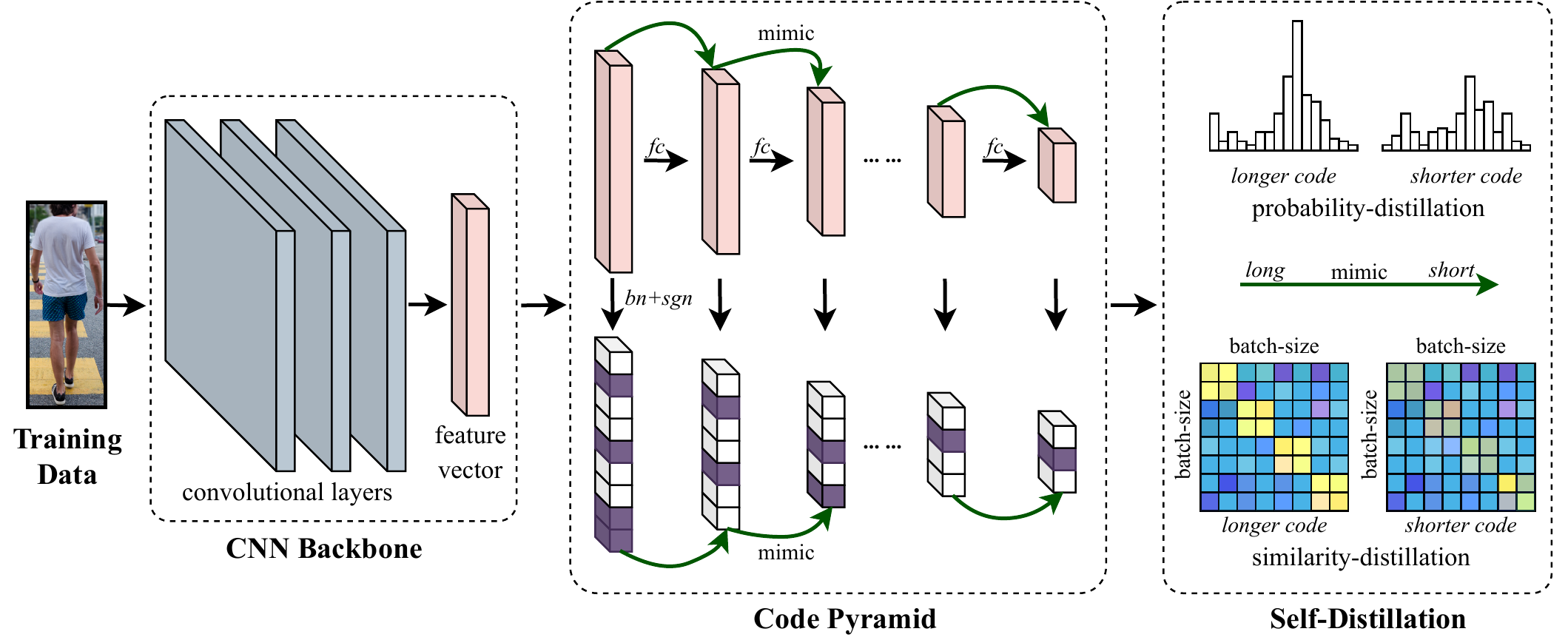}
\caption{All-in-One framework. It learns and enhances multiple codes of different lengths in a single framework with a code pyramid structure and self-distillation learning.
}
\label{figure:aio-framework}
\end{figure}

The All-in-One (AiO) framework aims to simultaneously learn and enhance multiple codes of different lengths in a single model, whose architecture can be seen in Figure \ref{figure:aio-framework}.
Specifically, it first utilizes a convolutional network to extract the real-value feature vectors, then learns multiple codes of different lengths in a pyramid structure, finally enhances the codes by encouraging shorter codes mimic longer codes via self-distillation.

\noindent \textbf{Learn Multiple Codes in a Pyramid Structure.}
The code pyramid learns multiple codes of different lengths, where the shorter codes are based on the longer codes. With such a structure, we can not only learn many codes in one shot, but also share the knowledge of longer codes with shorter codes. The equations are as below:
\begin{equation}
    v_{0} = F(x), \ \ 
    v_{k} = FC_{k}(v_{k-1}), \ \ k \in {1, 2, ..., N}, 
    \label{eq:hash-func}
\end{equation}
where $x$ is input image, $F$ is the CNN backbone, $N$ is the code number, $V = \{v_k\}_{k=1}^{N}$ are the real-value feature vectors with different lengths $L = \{l_k\}_{k=1}^{N}$, $FC_k$ is the fully-connected layers with $l_{k-1}$ input- and $l_{k}$ output-sizes.
After getting real-value features of different lengths, we can obtain their binary codes $B = \{b_{k}\}_{k=1}^{N}$ in the following equation.
\begin{equation}
    b_{k} = sgn(bn(v_{k})),
    \label{eq:quantization-func}
\end{equation}
where $bn$ is the batch normalization layer, $sgn$ is the symbolic function. 
We use the batch normalization layer because it normalizes the real-value features to be symmetric to $0$ and reduces the quantization loss.

\noindent \textbf{Enhance Codes with Self-Distillation Learning.}
As we discussed in the introduction section, the coarse ranking must be accurate enough to minimise missing true-match candidates in fine-grained ranking. Inspired by \cite{hinton2015distilling,tung2019similarity}, we introduce self-distillation learning to enhance the multiple codes in a single framework without importing extra teacher network.
Different from conventional distillation models, which imports an extra large teacher network to supervise a small student network, we perform distillation learning in a single network and achieve better performance, which is important for fast ReID.

Specifically, our self-distillation learning is composed of a probability- and a similarity- distillation. The probability-distillation transfers the instance-level knowledge in a from of softened class scores. Its formulation is given by
\begin{equation}
    \mathcal{L}_{pro} = \frac{1}{N-1} \sum_{k=1} ^{N-1} \mathcal{L}_{ce}(\sigma(\frac{z_{k+1}}{T}), \sigma(\frac{\hat{z}_{k}}{T})),
    \label{eq:prob-loss}
\end{equation}
where $\mathcal{L}_{ce}(\cdot, \cdot)$ denotes the cross-entropy loss, $\sigma$ is the softmax function, $\hat{z}_{k}/z_{k+1}$ means the output logits of the binary code $b_{k}/b_{k+1}$, $\hat{z}_{k}$ means it act as a teacher and fixed during training, $T$ is a temperature hyperparameter, which is set $1.0$ empirically.
The similarity-distillation transfers the knowledge of relationship from longer codes to shorter one, whose formulation is in Eq.(\ref{eq:sim-loss}). This is motivated by that as an image search task, ReID features should also focus on the relationship among samples, \textit{i.e.} to what extent the sample A is similar/dissimilar to sample B.
\begin{equation}
    \mathcal{L}_{sim} = \frac{1}{N-1} \sum_{k=1}^{N-1} \sum _{i,j}  ||\frac{1}{l_{k+1}}{G}^{i,j}_{k+1} - \frac{1}{l_{k}}  \hat{G}^{i,j}_{k}||^2,
    \label{eq:sim-loss}
\end{equation}
where $G^{i,j}_{k}/G^{i,j}_{k+1}$ is the Hamming distance between $b^{i}_{k}/b^{i}_{k+1}$ and $b^j_{k}/b^j_{k+1}$, $b^{i/j}_{k/k+1}$ is the binary code of image $x_i/x_j$ with length $l_{k}/l_{k+1}$, the $\hat{G}$ means that $G$ acts as a label and is fixed during the optimization process, thus contributes nothing to the gradients.

\noindent\textbf{Overall Objective Function and Training.}
Recent progresses on ReID have shown the effectiveness of the classification \cite{zheng2016person} and triplet \cite{hermans2017defense} losses. 
Thus, our final objective function includes our proposed probability- and similarity- distillation losses together with the classification and triplet losses as the final objective function. The formulation can be found in Eq.(\ref{eq:over-all-loss}),
\begin{equation}
    \mathcal{L} = \mathcal{L}_{ce} + \mathcal{L}_{tri} + \lambda_1 \mathcal{L}_{prob} + \lambda_2 \mathcal{L}_{sim}
    \label{eq:over-all-loss}
\end{equation}
Considering that the mapping function $sgn$ in Eq.(\ref{eq:quantization-func}) is discrete and Hamming distance in Eq.(\ref{eq:quantization-func}) is not differentiable, a natural relaxation \cite{Li2016Feature} is utilised in Eq.(\ref{eq:over-all-loss}) by replacing $sgn$ with $tanh$ and changing the Hamming distance to the inner-product distance. Finally, our All-in-One framework can be optimized in an end-to-end way by minimizing the loss in Eq.(\ref{eq:over-all-loss}).

\subsection{Distance Threshold Optimization}

\begin{algorithm}[t]
\caption{Distance Threshold Optimization}
\hspace*{0.02in} {\bf Input:}
Trained Model in Eq.(\ref{eq:quantization-func}), Validation Data $(X_v, Y_v)$ \\
\hspace*{0.02in} {\bf Output:}
Thresholds $\{T_{i}\}_{i=2}^{N}$
\begin{algorithmic}[1]
\For{$k = \{1, 2, ..., n-1\}$}
    \State $B_{k}$: Extract binary codes of validation set with length $l_k$ via Eq.(\ref{eq:quantization-func})
    \State $D^{r}$: Hamming distances of relevant pairs $(b^i_{k}, b^j_{k})$, where $y^i=y^j$
    \State $D^{n}$: Hamming distances of non-relevant pairs $(b^i_{k-1}, b^j_{k-1})$, where $y^i \not= y^j$
    \State $PDF^r, PDF^n$: Probability distribution function of $D^{r}$ and $D^{n}$ of in Eq.(\ref{eq:gaussian})
    \State $CDF^r, CDF^n$: Cumulative Distribution Function of $D^{r}$ and $D^{n}$ in Eq.(\ref{eq:gaussian})
    \State $t_{n+1}$: Maximize $F_\beta$ score in Eq.(\ref{eq:f-beta}) and return $t_{n+1}$
\EndFor
\State return $T=\{t_i\}_{i=2}^N$
\end{algorithmic}
\end{algorithm}

After getting the multiple codes of different lengths $B=\{b_i\}_{i=1}^{N}$, we can perform the Coarse-to-Fine (CtF) search.
There are two tips in CtF search, \textit{i.e.} high accuracy and fast speed.
For fast speed, the candidate number returned by coarse search should be small. For high accuracy, the candidates returned by coarse search should include relevant images as more as possible. 
But the two requirements are naturally conflicting. Thus, it is important to find the proper thresholds to optimally balance the two targets, \textit{i.e.} both high accuracy and fast speed.
One simple solution is brute search via cross-validation. However, the search space is too large. For example, if we have multiple binary codes of lengths $L = \{32, 128, 512, 2048\}$, the complexity of the brute search will be $\prod_{L} > 4 \times 10^{9}$ times.

In this part, we propose a novel Distance Threshold Optimization (DTO) algorithm which solves the time-consuming brute parameter search task with a simple optimization process. 
Specifically, inspired by \cite{goutte2005probabilistic}, we first explicitly formulate the two sub-targets as two scores in Eq.(\ref{eq:p-r-f}), \textit{i.e.} precision ($P$) and recall ($R$) scores. 
Then we balance the two sub-targets by mixing the two scores with a single parameter $\beta$ and get $F_{\beta}$ score in Eq.(\ref{eq:p-r-f}). 
\begin{equation}
\begin{split}
    P = \frac{TP}{TP+FP}, \ \ R = \frac{TP}{TP+FN} , \ \ F_{\beta} = (\beta^2+1)\frac{PR}{\beta^2P+R}
\end{split}
\label{eq:p-r-f}
\end{equation}
Here, $TP$ is the number of relevant images in the candidates, $FP$ is the number of non-relevant images in the candidates and $FN$ is not retrieved relevant samples.
As we can see, the precision score $P$ means the rate of relevant images in the candidates. Usually a high $P$ means a small candidate number, which is good for fast speed.
The recall score $R$ represents the rate of returned relevant samples in the total relevant samples. A high $R$ score means more returned relevant samples, which is important for high accuracy.
The $F_{\beta}$ mixed the precision and recall scores with a parameter $\beta$, which considers both speed and accuracy.
\begin{equation}
\begin{split}
        PDF(t) = \frac{1}{\sigma\sqrt{2\pi}} exp({-\frac{(t-u)^2}{\sigma\sqrt{2}}}), \ \ 
        CDF(t) = \frac{1}{2}(1 + erf\frac{t-u}{\sigma\sqrt{2}})
\end{split}
\label{eq:gaussian}
\end{equation}
\begin{equation}
F_{\beta} = \frac{CDF^{r}(\beta^2 + 1)}{CDF^{n} + CDF^{r} + \beta^2(1-CDF^{n} + CDF^{r})}
\label{eq:f-beta}
\end{equation}
Considering that TP/FP/FN are statistics which cannot be optimized, we replace them with two Gaussian cumulative distribution functions in form of Eq.(\ref{eq:gaussian}) (right), whose parameters $u$ and $\sigma$ are estimated by fitting a validation set using the Gaussian probability distribution function in Eq.(\ref{eq:gaussian}) (left). 
Finally, by maximizing the $F_{\beta}$ in Eq.(\ref{eq:f-beta}), we can get the optimal distance thresholds $T=\{t_k\}_{k=2}^{N}$ balanced by $\beta$.

\iffalse
\subsection{Discussion}

\noindent\textbf{Different with Distillation Model.}
%
Model distillation \cite{hinton2015distilling} requires an extra large teacher network to guide a small student one. Different from it, our self-distillation encourages shorter codes to mimic longer codes in a single network, which is more efficient. 

\noindent\textbf{Different with $F_{\beta}$ score.}
%
Precision, recall and F scores are evaluation protocols, which need to
\fi

\section{Experiments}

\subsection{Dataset and Evaluation Protocols}

% \begin{table}[ht!]
% \begin{center}
% \small
% \label{tab:datasets}
% \scalebox{1.0}{
% \begin{tabular}{c|c|c|c}
% \hline
% \hline
% \multicolumn{1}{c|}{\multirow{2}{*}{Dataset}} & \multicolumn{1}{c|}{\multirow{2}{*}{\begin{tabular}[c]{@{}c@{}}Train Nums\\ (ID/Image)\end{tabular}}} & \multicolumn{2}{c}{Testing Nums (ID/Image)}                \\ \cline{3-4} 
% \multicolumn{1}{c|}{}                         & \multicolumn{1}{c|}{}                                                                                   & \multicolumn{1}{c|}{Gallery} & \multicolumn{1}{c}{Query} \\ \hline
% Market-1501       & 751/12,936 & 750/19,732 & 750/3,368 \\
% Market-1501+500k  & 751/12,936 & 750/519,732 & 750/3,368 \\
% DukeMTMC-reID     & 702/16,522 & 1,110/17,661 & 702/2,228 \\
% \hline
% \hline
% \end{tabular}
% }
% \end{center}
% \caption{Dataset details. We evaluate our method
%   on 3 public ReID datasets, including Market-1501,
%   DukeMTMC-reID and one large-scale one Market-1501+500k.}
% \end{table}

\noindent\textbf{Datasets.}
We extensively evaluate our proposed method on two common datasets (Market-1501 \cite{zheng2015scalable} and DukeMTMC-reID \cite{zheng2017unlabeled}) and one large-scale dataset (Market-1501+500k\cite{zheng2015scalable}).
The Market-1501 dataset contains 1,501 identities observed under 6 cameras, which are splited into 12,936 training, 3,368 query and 15,913 gallery images.
The Market-1501+500k enlarges the gallery of Market-1501 with extra
500,000 distractors, making it more challenging for both accuracy and speed.
DukeMTMC-reID contains 1,404 identities with 16,5522 training, 2,228 query and 17,661 gallery images.

\noindent\textbf{Evaluation Protocols.}
For accuracy, we use standard metrics including Cumulative Matching Characteristic (CMC) curves and mean average precision (mAP). All the results are from a single query setting.
To evaluate speed, we use average query time per image, including distance computation and sorting time. 
For fair evaluation, we do not use any parallel algorithm for distance computation and sorting.

\subsection{Implementation Details}

We implemented our method with Pytorch on a PC with 2.6Ghz Intel Core i5 CPUs, 10GB memory, and a NVIDIA RTX 2080Ti GPU. 
For a fair comparison and following \cite{luo2019bag,liu2019adversarial}, we use ResNet50 \cite{he2016deep} as the CNN backbone.
In training stage, each image is resized to $256\times128$ and augmented by horizontal flip and random erasing
\cite{zhong2017random}. 
A batch data includes 64 images from 16 different persons, where every person includes 4 images.
The lengths $L=\{l_k\}_{k=1}^{N}$ of multiple codes are empirically set $\{32, 128, 512, 2048\}$. The margin in the triplet loss in Eq.(\ref{eq:over-all-loss}) is 0.3.
The framework is optimized by Adam \cite{kingma2014adam} with total epochs 120.
Its initial learning rate is 0.00035, which is warmed up for 10 epochs and decayed to its $0.1\times$ and $0.01\times$ at 40 and 70 epochs. 
We randomly split the training data into a training and a validation set according to $6:4$, then decide the parameters via cross-validation, After that, we train our method with all training data. 
$\lambda_1$ and $\lambda_2$ in Eq.(\ref{eq:over-all-loss}) are set as 1.0 and 1,000, and $\beta$ in Eq.(\ref{eq:f-beta}) is set 2.0. The three paramters are decided via cross validation.
Code is available at github\footnote{\url{https://github.com/wangguanan/light-reid}}.

\subsection{Comparisons with Non-Hashing ReID Methods}

\begin{table}[t]
\begin{center}
\small
\begin{tabular}{l|cc|ccc|ccc}
\hline
\hline 
\multicolumn{1}{l|}{\multirow{2}{*}{Methods}} & \multicolumn{2}{c|}{Code} & \multicolumn{3}{c|}{Market-1501} & \multicolumn{3}{c}{DukeMTMC-reID} \\
\multicolumn{1}{c|}{} & \multicolumn{1}{c}{Type} & \multicolumn{1}{c|}{Length}                         
& \multicolumn{1}{c}{R1(\%)} & \multicolumn{1}{c}{mAP(\%)} & \multicolumn{1}{c|}{\textbf{\textit{Q.Time}}(s)} 
& \multicolumn{1}{c}{R1(\%)} & \multicolumn{1}{c}{mAP(\%)} & \multicolumn{1}{c}{\textbf{\textit{Q.Time}}(s)} \\
\hline
 PSE \cite{sarfraz2018a}            & $\mathbf{R}$  & 1,536  & 78.7 & 56.0 & - & -   & - & -     \\
 PN-GAN \cite{qian2018pose}         & $\mathbf{R}$  & 1,024  & 89.4 & 72.6 & - & 73.6 & 53.2 & - \\
 % HA-CNN [???]                      & $\mathbf{R}$  & 1024  & 91.2 & 75.7 & - & 80.5 & 63.8 & - \\
 IDE \cite{zheng2016person}         & $\mathbf{R}$  & 2,048  & 88.1 & 72.8 & - & 69.4 & 55.4 & - \\
 Camstyle \cite{zhong2018camera}    & $\mathbf{R}$  & 2,048  & 88.1 & 68.7 & - & 75.3 & 53.5 & - \\
 PIE \cite{zheng2019pose}           & $\mathbf{R}$  & 2,062  & 87.7 & 69.0 & - & 79.8 & 62.0 & - \\
 % DuATM [???]                        & $\mathbf{R}$  & -     & 91.4 & 76.6 & - & 81.2 & 62.3 & - \\
 % Mancs [???]                        & $\mathbf{R}$  & 2048  & 93.1 & 82.3 & - & 84.9 & 71.8 & - \\
 % SVD [???]                          & $\mathbf{R}$  & -     & 82.3 & 62.1 & - & 76.7 & 56.8 & - \\
 % TriNet [???]                       & $\mathbf{R}$  & -     & 84.9 & 69.1 & - & - & - & -  \\
 BoT~\cite{luo2019bag}              & $\mathbf{R}$  & 2,048  & \underline{94.1} & \underline{85.7} & \underline{$2.2\times10^0$} & \underline{86.4} & \underline{76.4} & \underline{$2.0\times10^{0}$} \\
\hline
 SPReID \cite{kalayeh2018human}    & $\mathbf{R}$  & 10,240  & 92.5 & 81.3 & - & 84.4 & 71.0 & - \\
 PCB \cite{sun2018beyond}          & $\mathbf{R}$  & 12,288  & 93.8 & 81.6 & $6.9\times10^0$ & 83.3 & 69.2 & $6.3\times10^{0}$ \\
 VPM \cite{sun2019perceive}        & $\mathbf{R}$  & 14,336  & 93.0 & 80.8 & - & 83.6 & 72.6 & - \\  
%  Pyramid [???]                      & $\mathbf{R}$  & 2048  & 92.8 & 82.1 & - & - & - & - \\
\hline
\textbf{CtF (ours)}                                & $\mathbf{B}$  & 2,048  & \textbf{93.7} & \textbf{84.9} & $\mathbf{4.6\times10^{-2}}$ & \textbf{87.6} & \textbf{74.8} & \textbf{$\mathbf{3.7\times10^{-2}}$} \\
\hline
\hline
\end{tabular}
\end{center}
\caption{Comparisons with non-hashing ReID methods using real-value
  features of different lengths on Market-1501 and
  DukeTMTC-reID. $\mathbf{B}$: binary code, $\mathbf{R}$:
  real-value feature. Longer real-value features have higher
  accuracy but slower query speed. Our model CtF (including AiO) has very fast query
  speed (two orders of magnitude faster) and comparable accuracy with
  non-hashing ReID methods. }
\label{tab:sota-non-hashing}
\end{table}

Non-hashing ReID use longer real-value features, such as
2048-dimensional $float64$ features, for a better accuracy. This significantly affects their speed, \textit{i.e.} query time.
Table~\ref{tab:sota-non-hashing} shows that our proposed CtF
(including AiO) method is
significantly faster than non-hashing ReID methods (two orders of
magnitude). CtF also achieves
very competitive accuracy with close Rank-1 (93.7\% vs. 94.1\%) and mAP (87.6\% vs. 86.4\%) scores
of the best non-hashing ReID mehtod BoT \cite{luo2019bag}  on Market-1501 and DukeMTMC-reID, and better than all the other non-hashing
methods using different feature length, of which 5 methods have features shorter than 2,062 (PSE \cite{sarfraz2018a}, IDE \cite{zheng2016person}, PN-GAN \cite{qian2018pose}, CamStyle \cite{zhong2018camera}, PIE \cite{zheng2019pose})
and 3 methods have features longer than 10,240 (SPReID \cite{kalayeh2018human}, PCB \cite{sun2018beyond}, VPM \cite{sun2019perceive}).
Overall, longer feature usually contributes to higher accuracy but
with slower speed. For example, SPReID, PCB and VPM take features
longer than 10,240 and achieves $92\%$-$93\%$ and $83\%$-$84\%$ Rank-1
scores on Market-1501 and DukeMTMC-reID datasets, respectively. The
others utilize features no longer than 2,048 achieving Rank-1 score less than $92\%$ and $80\%$.
On the other hand, the query speed of those methods with long features is much slower. For example, 
PCB takes $6.9s$ and $6.3s$ for query each image on the two datasets
respectively. This is $3$-$4\times$ slower than IDE with $2s$ on either dataset. 
Specifically, CtF performs much faster than non-hashing methods and
significantly, it achieves much better accuracy than comparable
length real-value feature model. For example, CtF achieves
$93.7\%/87.6\%$ Rank-1 scores on Market-1501/DukeMTMC-reID, as
compared to BoT having $94.1\%/86.4\%$ respectively.
This is because CtF (including AiO) utilizes all-in-one framework together with coarse-to-fine search strategy, which not only learns powerful binary code, but also complementarily uses short and long codes for both high accuracy and fast speed.

\subsection{Comparisons with Hashing ReID Methods}

\begin{table}[t]
\begin{center}
\begin{tabular}{c|c|ccc|ccc}
\hline
\hline 
\multicolumn{1}{c|}{\multirow{2}{*}{Methods}} & \multicolumn{1}{c|}{Code} & \multicolumn{3}{c|}{Market-1501} & \multicolumn{3}{c}{DukeMTMC-reID} \\
\multicolumn{1}{c|}{} & \multicolumn{1}{c|}{Length} & 
\multicolumn{1}{c}{R1(\%)} & \multicolumn{1}{c}{mAP(\%)} & \multicolumn{1}{c|}{\textbf{Q.Time}(s)} &
\multicolumn{1}{c}{R1(\%)} & \multicolumn{1}{c}{mAP(\%)} & \multicolumn{1}{c}{\textbf{Q.Time}(s)} \\
\hline
DRSCH \cite{Zhang2015Bit}       & 512   & 17.1 & 11.5 & - & 19.3 & 13.6 & - \\
DSRH \cite{zhao2015deep}        & 512   & 27.1 & 17.7 & - & 25.6 & 18.6 & - \\
HashNet \cite{cao2017hashnet}   & 512   & 29.2 & 19.1 & - & 40.8 & 28.6 & - \\
DCH \cite{cao2018deep}          & 512   & 40.7 & 20.2 & - & 57.4 & 37.3 & - \\
CSBT \cite{chen2017fast}        & 512   & 42.9 & 20.3 & - & 47.2 & 33.1 & - \\
PDH \cite{zhu2017part}          & 512   & 44.6 & 24.3 & - & - & - & - \\
DeepSSH \cite{zhao2018deepssh}  & 512   & 46.5 & 24.1 & - & - & - & - \\
ABC \cite{liu2019adversarial}   & 512   & 69.4 & 48.5 & $9.8\times10^{-2}$ & 69.9 & 52.6 & $7.5\times10^{-2}$ \\
ABC \cite{liu2019adversarial}   & 2,048  & 81.4 & 64.7 & $2.8\times10^{-1}$ & 82.5 & 61.2 & $2.0\times10^{-1}$ \\
\hline
\multirow{5}{*}{\textbf{CtF (ours)}} & AiO+32  & 60.0 & 37.7 & \underline{$3.4\times10^{-2}$} & 49.5 & 28.7 & \underline{$2.3\times10^{-2}$} \\
                            & AiO+128 & 88.9 & 71.0 & $4.2\times10^{-2}$ & 78.6 & 59.4 & $3.2\times10^{-2}$\\
                            & AiO+512 & 92.8 & 82.2 & $9.8\times10^{-2}$ & 85.4 & 71.6 & $7.5\times10^{-2}$\\
                            & AiO+2,048 & \underline{93.7} & \underline{85.4} & $2.8\times10^{-1}$ & \underline{87.7} & \underline{75.7} & $2.0\times10^{-1}$ \\
                            & AiO+CtF & \textbf{93.7} & \textbf{84.0} & $\mathbf{4.6\times10^{-2}}$ & \textbf{87.6} & \textbf{74.8} & $\mathbf{3.7\times10^{-2}}$ \\
\hline
\hline
\end{tabular}
\end{center}
\caption{Comparisons with state-of-the-art hashing ReID methods on
  Market-1501 and DukeTMTC-reID. AiO+$k$ means learning multiple codes
  with all-in-one framework, but querying with only the code of length
  $l_{k}$. Aio+CtF not only learns multiple codes with all-in-one
  framework, but also query with coarse-to-fine search strategy. Our
  AiO+CtF achieve a good balance between accuracy and speed.} 
\label{tab:sota-hashing}
\end{table}

Hashing ReID methods learn binary codes using a hashing algorithm. 
Binary codes are good for speed but sacrifice model accuracy. To mitigate
this problem, the state-of-the-art hashing ReID 
methods usually employ long codes such as $2048$. In binary coding,
$2048$ is relatively very long as compared to the more commonly used
$512$ length, unlike in real-value feature length compared above.  
Table \ref{tab:sota-hashing} shows that CtF (with AiO) not only
achieves the best accuracy (even compared to much shorter code length
used by other hashing methods), but also is significantly faster than
existing hashing ReID methods (even compared to the same code length
used by other hashing methods).
Overall, hashing ReID methods usually perform much worse than
non-hashing methods. For example, best non-hashing ReID methods
achieves $94.1\%$ and $86.4\%$ Rank-1 scores on Market-1501 and
DukeMTMC-reID respectively. But the best hashing ReID method only obtains $81.4\%$ and
$82.5\%$ Rank-1 scores.
Moreover, existing hashing ReID models can increase accuracy by using
longer code length and compromising speed. For example, ABC with
512-dimensional binary codes achieves $69.4\%/69.9\%$ Rank-1 scores and
$9.8/7.5\times10^{-2}s$ query time per probe image. When using 2048
binary codes, its Rank-1 scores increase to $81.4\%/82.5\%$ with query
time slow down to $2.8/2.0\times10^{-1}s$. This observation is also verified with our method
CtF (with AiO) using different code lengths. 
Importantly, our method CtF (with AiO) significantly outperforms all
existing hashing ReID methods in terms of both accuracy (R1 12.3\% or
5.1\% better) and speed ($5\times$ faster).
Specifically, CtF with AiO achieves high accuracy very close to AiO
without CtF using 2048 code length, but yields significant speed
advantage that is comparable to much shorter 128 binary code
length. CtF obtains $93.7\%$ and $87.6\%$ Rank-1 scores, similar to AiO
without CtF of a fixed 2048 length at $93.7\%$ and
$87.7\%$.
%
%Overall, the experimental results show our effectiveness on both accuracy and speed.

\subsection{Evaluation on Large-Scale ReID}
\begin{figure}[t!]
\center
\includegraphics[width=\linewidth]{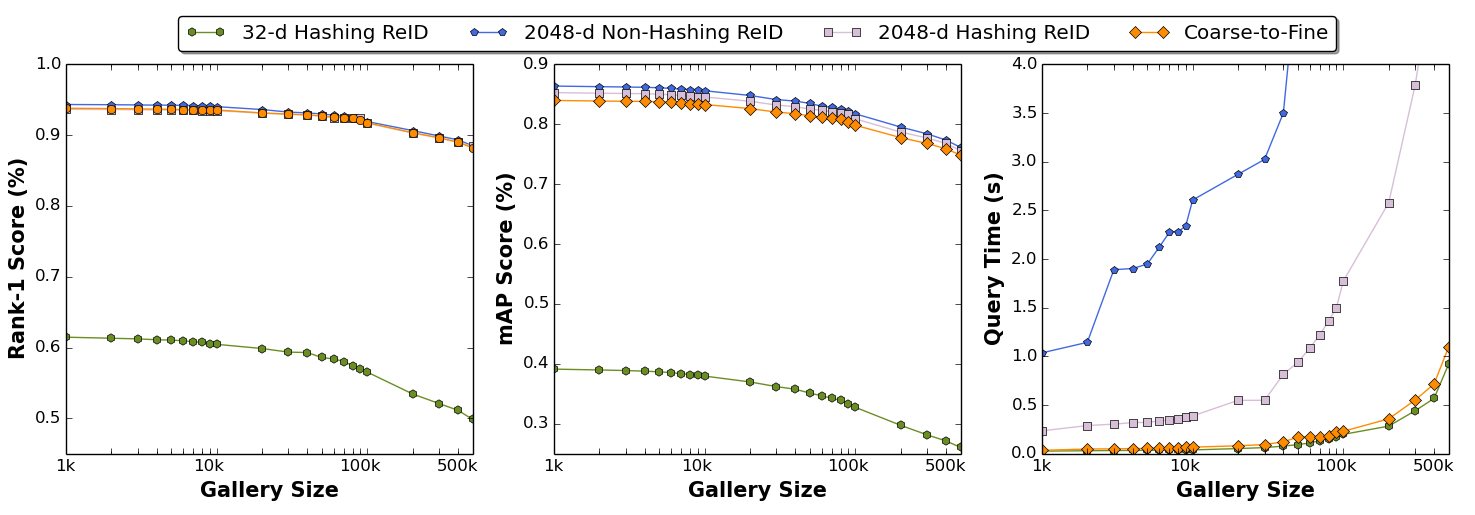}
\caption{Experimental results on large-scale ReID dataset Market-1501+500k. Our Coarse-to-Fine (CtF) get a high accuracy comparable with non-hashing ReID method of long code and fast speed comparable with hashing ReID method of short code.}
\label{fig:large-scale}
\end{figure}

Gallery size affects significantly ReID search accuracy and speed. To show the
effectiveness of our proposed Coarse-to-Fine (CtF) search strategy, we
evaluated it on a large-scale ReID dataset Market1501+500k. The
dataset is based on the Market-1501 and enlarged with $500,000$
distractors. 
%
% We compare our CtF with three ReID methods, including a non-hashing ReID method with 2048-dimensional real-value features, a hashing ReID model with long binary codes of 2048-dimension, and a hashing ReID model with short binary codes of 32-dimension. 
%
The experimental results are shown in Figure \ref{fig:large-scale}. We can observe the following phenomenons.

Firstly, with the increase of gallery size, for all methods, the
Rank-1 and mAP scores decrease, and the ReID speed per probe image 
slows down gradually. The reason is that more gallery images is more likely to contain
more difficult samples. They make ReID search more challenging. Also, the extra
gallery images significantly increase the time for computing all the
distance comparisons and sorting required for ReID each probe image.
Secondly, the non-hashing method with 2048-D real-value feature
achieves the best accuracy but the worst time. This is because the
real-value feature is more discriminative but slow to compute and sort.
Thirdly, for hashing ReID methods, the 2048-D binary code obtains
comparable ReID accuracy to that of the non-hashing model, but
$10\times$ faster. This is because Hamming distances and counting
sort are faster to compute. ReID speed of 32-D binary code is
$5\times$ faster than that of 2048-D binary codes, but its
accuracy drops dramatically. 
Finally, the proposed CtF model achieves a comparable accuracy to that
of the non-hashing method but the advantage of similar speed to that
of a hashing ReID method of 32-D binary code. Critically, the
advantage is independent of the gallery size.
Overall, these experiments demonstrate the effectiveness of CtF for
a large-scale ReID task.

\subsection{Model Analysis}

%
\iffalse
\begin{table}[t]
\begin{center}
\scalebox{1}{
\begin{tabular}{ccc|c|ccccc|ccccc}
\hline
\hline 
\multirow{2}{*}{CP} & \multirow{2}{*}{PD} & \multirow{2}{*}{SD} & \multirow{1}{*}{Code} & \multicolumn{4}{c|}{Rank-1(\%)} & \multicolumn{4}{c}{mAP(\%)} \\
 & & & Type &
%\multicolumn{1}{c}{8} &
\multicolumn{1}{c}{32} & 
\multicolumn{1}{c}{128} &
\multicolumn{1}{c}{512} & 
\multicolumn{1}{c}{2048} &
%\multicolumn{1}{|c}{8} &
\multicolumn{1}{c}{32} & 
\multicolumn{1}{c}{128} &
\multicolumn{1}{c}{512} & 
\multicolumn{1}{c}{2048} &\\ 
\hline 
\times      & \times        & \times        & $\mathbf{B}$    & 25.5 & 84.8 & 92.3 & 93.8 & 33.9 & 67.5 & 81.4 & 84.0\\
\Checkmark  & \times        & \times        & $\mathbf{B}$    & 54.4 & 87.8 & 92.7 & 93.8 & 35.0 & 72.2 & 81.7 & 84.8\\
\Checkmark  & \Checkmark    & \times        & $\mathbf{B}$    & 54.4 & 88.4 & 92.4 & 94.0 & 34.3 & 71.1 & 81.4 & 85.1\\
\Checkmark  & \times        & \Checkmark    & $\mathbf{B}$    & 60.5 & 88.4 & 92.9 & 94.0 & 37.0 & 71.6 & 82.3 & 85.4\\
\Checkmark  & \Checkmark    & \Checkmark    & $\mathbf{B}$    & 60.0 & 88.9 & 92.9 & 93.8 & 37.7 & 71.0 & 82.0 & 85.3 \\
\hline
\multicolumn{3}{c|}{upper bound}            & $\mathbf{R}$    & 82.7 & 90.9 & 93.4 & 94.2 & 66.7 & 78.9 & 84.3 & 85.4\\
\hline
\hline
\end{tabular}
}
\end{center}
\caption{Analysis of All-in-One (AiO) framework. $\mathbf{B}$ and $\mathbf{R}$ mean binary codes and real-valure features, respectively.}
\label{tab:ablation-study}
\end{table}
\fi

\begin{table}[t]
\begin{center}
\begin{tabular}{ccc|c|ccccc|ccccc}
\hline
\hline 
\multirow{2}{*}{\textbf{AiO}} & \multirow{2}{*}{\textbf{CP}} & \multirow{2}{*}{\textbf{SD}} & \multirow{1}{*}{Feature} & \multicolumn{5}{c|}{Rank-1(\%)} & \multicolumn{5}{c}{mAP(\%)} \\
 & & & Type &
%\multicolumn{1}{c}{8} &
\multicolumn{1}{c}{\ 32 \  } & 
\multicolumn{1}{c}{\ 128 \ } &
\multicolumn{1}{c}{\ 512 \ } & 
\multicolumn{1}{c}{2048} &
\multicolumn{1}{c|}{\ CtF \ } &
%\multicolumn{1}{|c}{8} &
\multicolumn{1}{c}{\ 32 \ } & 
\multicolumn{1}{c}{\ 128 \ } &
\multicolumn{1}{c}{\ 512 \ } & 
\multicolumn{1}{c}{2048} &
\multicolumn{1}{c}{\ CtF \ } \\
\hline 
$\times$    & $\times$    & $\times$     & $\mathbf{B}$   & - & - & - & - & - & - & - & - & - &  \\
\Checkmark  & $\times$        & $\times$      & $\mathbf{B}$    & 25.5 & 84.8 & 92.3 & 93.8 & 92.5 & 33.9 & 67.5 & 81.4 & 85.3 & 75.1 \\
\Checkmark    & \Checkmark    & $\times$        & $\mathbf{B}$    & 54.4 & 87.8 & 92.7 & 93.8 & 93.0 & 35.0 & 72.2 & 81.7 & 85.3 & 80.2 \\
\Checkmark    & \Checkmark    & \Checkmark    & $\mathbf{B}$      & 60.0 & 88.9 & 92.9 & 93.8 & 93.7 & 37.7 & 71.0 & 82.0 & 85.3 & 84.0 \\
\hline
\multicolumn{3}{c|}{upper bound}            & $\mathbf{R}$    & 82.7 & 90.9 & 93.4 & 94.2 & - & 66.7 & 78.9 & 84.3 & 85.4& - \\
\hline
\hline
\end{tabular}
\end{center}
\caption{Analysis of the All-in-One (AiO) framework. \textbf{CP}: learn multiple codes in a pyramid structure, otherwise separate models. \textbf{SD}: enhance binary codes via self-distillation. $\mathbf{B}$ and $\mathbf{R}$ mean binary codes and real-value features, respectively.} 
\label{tab:analysis-aio}
\end{table}

\noindent \textbf{Analysis of AiO.}
The All-in-One (AiO) framework aims to learn and enhance multiple
codes of different lengths in a single model. It uses code pyramid
(CP) structure and self-distillation (SD) learning. 
%To validate its effectiveness, we evaluated AiO by analyzing its CP and SD components explicitly, and reporting the performance of querying with single code as compared to coarse-to-fine (CtF) strategy. 
Results are in Table \ref{tab:analysis-aio}. 
Firstly, longer codes contribute to better accuracy. This can be seen in all settings no matter whether CP or SD is used and what code type is.
Secondly, when using short codes, real-value features is much better than binary ones. But for long codes, they obtain similar accuracy.
For example, the 32-dimensional real-value feature obtains $82.7\%$
Rank-1 score, outperforming the 32-dimensional binary code by $60\%$,
where the latter achieved only $25.5\%$. But when using 2048 code length, binary
codes and real-valure features both achieve approx. Rank-1 $94\%$
and mAP $84\%$. This suggests that the quantization loss
of short codes is significantly worse than that of longer codes. 
Thirdly, learning with code pyramid (CP) structure or
self-distillation (SD) improves short codes
significantly. 
For example, CP+SD boosts the 32-dimensional binary codes from
$25.5\%$ to $60.0\%$ in Rank-1 score, upto $35\%$ gain. 
It is evident that both code pyramid (CP) structure and
self-distillation (SD) learning contribute to the effectiveness of the
coarse-to-fine (CtF) search strategy, and significantly improve model performance.

\begin{figure}[t]
\center
\includegraphics[scale=0.40]{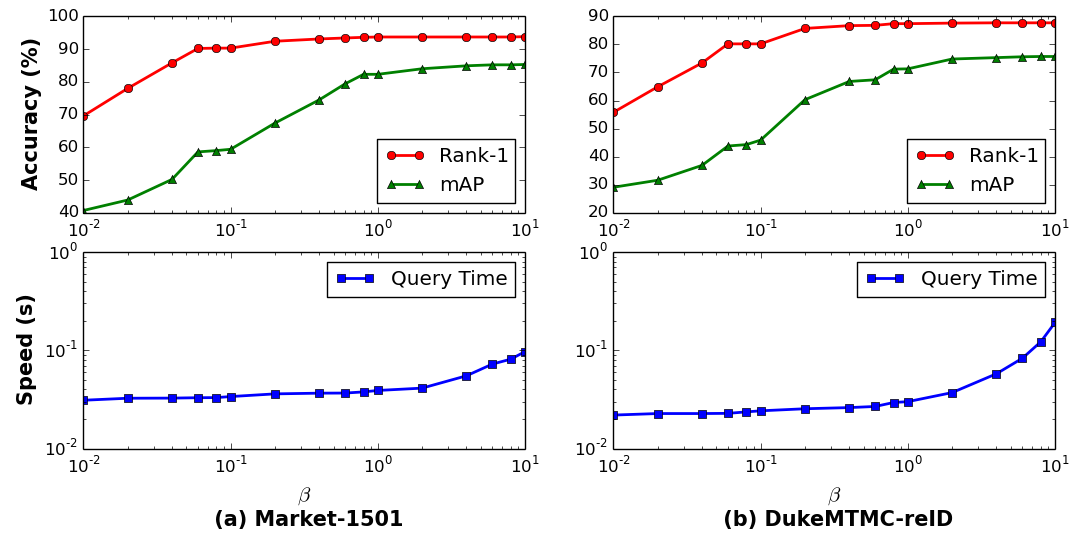}
\caption{Accuracy and Speed controlled by $\beta$. With the increase of $\beta$, the accuracy increases and speed becomes slow gradually.}
\label{fig:beta}
\end{figure}

\noindent\textbf{Analysis of DTO.}
We further analyzed parameter $\beta$ of the Distance Threshold
Optimization (DTO) algorithm, which controls the balance between ReID accuracy and speed. 
Figure \ref{fig:beta} show the model accuracy and speed using different
$\beta$ value on Market-1501 and DukeMTMC-reID.
Firstly, it is evident that the value of $\beta$ has a good control of
accuracy and speed, increasing $\beta$ slows down the speed but
improves accuracy. For example, when $\beta=10^{-2}$, ReID is fastest
at approx. 0.03 and 0.02 seconds to ReID each probe image on
Market-1501 and DukeMTMC-reID, but with mAP scores only at $40\%$ and $30\%$.
In contrast, $\beta=10^{1}$ gives high mAP $85\%$ and $75\%$, but the
query speed is $5\times$ slower at approx. $0.1$ and $0.2$ seconds.
Secondly, when $\beta$ is close to $10^{0}$, Rank-1 and mAP are almost
peaked with a good balance on speed. 

\section{Conclusion}

In this work, we proposed a novel Coarse-to-Fine (CtF) search strategy
for faster person re-identification whilst also improve accuracy on
conventional hashing ReID. 
% CtF first coarsely ranks a gallery using shorter binary codes, then iteratively utilises longer binary codes to further refine on ranking selected top candidates with increasing accuracy.  
% In order to implement the CtF strategy, a novel All-in-One (AiO)
% framework together with a Distance Threshold Optimization (DTO)
% algorithm are formulated.
% The former simultaneously learns and enhances multiple binary codes of
% different lengths in a single model. 
% The latter solves the complex parameter search task by a simple
% optimization process. The balance between search accuracy and
% speed is easily controlled by a single parameter. 
%
Extensive experiments show that our method is 
$5\times$ faster than existing hashing ReID methods but achieves comparable
accuracy with non-hashing ReID models that are 50$\times$
slower.

\section*{Acknowledgement}
This work was supported in part by the National Key R\&D Program of China (Grant 2018YFC2001700), by the National Natural Science Foundation of China (Grants 61720106012, and U1913601), by the Beijing Natural Science Foundation (Grants L172050), by the Strategic Priority Research Program of Chinese Academy of Sciences (Grant XDB32040000), by the Youth Innovation Promotion Association of CAS (2020140), the Alan Turing Institute Turing Fellowship, and Vision Semantics Ltd. 
  
\bibliographystyle{splncs04}
\bibliography{egbib}
\end{document}